\pdfoutput=1

\documentclass[11pt]{article}

\usepackage[preprint]{acl}

\usepackage{times}
\usepackage{latexsym}

\usepackage[T1]{fontenc}

\usepackage[utf8]{inputenc}

\usepackage{microtype}

\usepackage{inconsolata}

\usepackage{graphicx}
\usepackage{multirow}

\title{Do LLMs Have the Generalization Ability in Conducting Causal Inference?}

\author{Chen Wang\footnotemark[2]\ \ \ Dongming Zhao\footnotemark[4] \ \ \ Bo Wang\footnotemark[2]\footnotemark[3]\thanks{Corresponding author.}  \ \ \ Ruifang He\footnotemark[2] \ \ \ Yuexian Hou\footnotemark[2]\\
\footnotemark[2]College of Intelligence and Computing, Tianjin University, Tianjin, China\\ \footnotemark[3]Institute of Applied Psychology, Tianjin University, Tianjin, China\\ 
\footnotemark[4]China Mobile Communication Group Tianjin Co.,Ltd\\
\footnotemark[2]\footnotemark[3]\texttt{\{chen\_wang, bo\_wang, rfhe, yxhou\}@tju.edu.cn}, \footnotemark[4]\texttt{waitman\_840602@163.com}
}

\begin{document}
\maketitle
\begin{abstract}

In causal inference, generalization capability refers to the ability to conduct causal inference methods on new data to estimate the causal-effect between unknown phenomenon, which is crucial for expanding the boundaries of knowledge. Studies have evaluated the causal inference capabilities of Large Language Models (LLMs) concerning known phenomena, yet the generalization capabilities of LLMs concerning unseen phenomena remain unexplored.
In this paper, we selected four tasks: Causal Path Discovery (CP), Backdoor Adjustment (BA), Factual Inference (FI), and Counterfactual Inference (CI) as representatives of causal inference tasks. To generate evaluation questions about previously unseen phenomena in new data on the four tasks, we propose a benchmark generation framework, which employs randomly generated graphs and node names to formulate questions within hypothetical new causal scenarios. Based on this framework, we compile a benchmark dataset of varying levels of question complexity. We extensively tested the generalization capabilities of five leading LLMs across four tasks. Experiment results reveal that while LLMs exhibit good generalization performance in solving simple CP, FI, and complex CI questions, they encounter difficulties when tackling BA questions and face obvious performance fluctuations as the problem complexity changes. Furthermore, when the names of phenomena incorporate existing terms, even if these names are entirely novel, their generalization performance can still be hindered by interference from familiar terms. Our code and dataset are available at: \url{https://github.com/prayingsociety/CI_Bench}

\end{abstract}

\section{Introduction}
\label{sec:intro}
In scientific research, using causal inference to study the connections between unseen phenomena is extremely important \citep{zhang2023understandingcausalitylargelanguage}. This requires not only understanding causality but also possessing the generalization ability to handle the previously unseen questions. The recently emerged LLMs have once again become a hot research topic due to their powerful capabilities \citep{naveed2024comprehensiveoverviewlargelanguage}. Some studies have considered the potential of combining LLMs with causal inference (\citealp{liu2024largelanguagemodelscausal}, \citealp{kıcıman2024causalreasoninglargelanguage}, \citealp{feder-etal-2022-causal}).
In this background, understanding how well LLMs can solve causal inference problems becomes crucial. 
Moreover, since LLMs have vast knowledge reserves, sometimes their answers are merely reproductions of memorized information rather than conclusions derived through reasoning (\citealp{zečević2023causalparrotslargelanguage}, \citealp{yang2024critical}). Therefore, the generalization ability of LLMs when faced with unseen data is also worth considering. Although there have been many works that evaluated the performance of LLMs on causal reasoning, there is still a lack of works that assess the generalization ability of LLMs targeted at causal inference tasks. Therefore, we aimed to create a general benchmark generation framework to evaluate LLMs' abilities in solving causal inference tasks on unseen data.

To achieve this goal, we selected four tasks that are crucial for addressing the problem of causal inference: Causal Path Finding (CP), Backdoor Adjustment (BA), Factual Inference (FI) and Counterfactual Inference (CI). We assumed that performance on these important subtasks can represent the model's ability to solve causal inference problems. We will further discuss the reasons for selecting these tasks in Section~\ref{sec:method-taskselection}.
To construct a benchmark on these tasks, we proposed a benchmark generation framework that used a randomly generated causal graph and node names to create unseen causal scenarios. Based on this framework, we constructed a benchmark dataset with different question complexity on the four tasks.
We implemented extensive experiments with five representative LLMs on the constructed benchmark dataset using 7 different prompts in 3 types. Our experiments revealed that:
\begin{enumerate}
    \item Although LLMs showed good performance on simple CP questions, they still struggled with complex scenarios and exhibited significant performance differences when solving simple and complex problems.
    \item LLMs achieved relatively consistent cross-performance across varying complexities of the BA tasks with low average accuracy, indicating that they have not effectively mastered the ability to solve such problems.
    \item LLMs performed well on the FI task. While showing performance drop with the increase of complexity in FI, they demonstrated better performance on complex CI problems compared to simpler ones.
    \item When confronted with unseen phenomena in familiar domains, LLMs exhibited a consistent decline in performance.
    \item Overall, the experimental results indicate that LLMs possess a certain degree of generalization ability. However, they still cannot perform causal inference well independently in new research contexts.
\end{enumerate}

Our main contributions are:
\begin{itemize}
    \item Proposed a framework for generating a benchmark dataset to assess causal inference generalization across four tasks—Causal Path Finding (CP), Backdoor Adjustment (BA), Factual Inference(FI), and Counterfactual Inference (CI)—by constructing new unseen causal scenarios using randomly generated causal graphs and node names.
    \item Constructed an evaluation dataset with 4,188 problems of varying complexity using the proposed framework for four tasks and conducted extensive experiments on five representative LLMs using three types of prompts.
    \item Key conclusions drawn from the experiments:
    Despite strong performance in specific scenarios, LLMs' overall performance is unstable. They tend to associate unseen terms with existing commonsense knowledge. Overall, LLMs demonstrate limited generalization ability in causal inference tasks.
\end{itemize}

\section{Related Works}
\label{sec:relatedworks}
\textbf{Causal Reasoning evaluations of LLMs.}   There have been many research evaluated LLMs' causal reasoning abilities in several aspects including pair-wide causal relation identification (\citealp{frohberg-binder-2022-crass}, \citealp{gao-etal-2023-chatgpt}, \citealp{zečević2023causalparrotslargelanguage}, \citealp{kıcıman2024causalreasoninglargelanguage}), commonsense causal reasoning (\citealp{zhang-etal-2020-reasoning}, \citealp{zhang-etal-2020-winowhy}, \citealp{singh-etal-2021-com2sense}, \citealp{tu2023causaldiscoveryperformancechatgptcontext},  \citealp{zečević2023causalparrotslargelanguage}, \citealp{liu-etal-2023-magic}), and graph-based tasks (\citealp{jin2024largelanguagemodelsinfer}, \citealp{jin2024cladderassessingcausalreasoning}, \citealp{chen2024clearlanguagemodelsreally}). Both pair-wide causal relation identification and commonsense causal reasoning rely on knowledge of existing phenomena to perform causal reasoning. Still, this approach is ineffective when confronting entirely new and unseen phenomena.
Although \citet{jin2024cladderassessingcausalreasoning} has conducted a comprehensive evaluation on causal inference ability, their reasoning is more focused on numerical data, while ours emphasizes structure and logic-based reasoning. Besides, we investigated LLMs' causal inference generalization abilities on different question complexities while they have questions generated from the graph with up to 4 nodes, which is quite limited.
Recently, \citet{chen2024clearlanguagemodelsreally} conducted a comprehensive evaluation on causal graph understanding ability. They focused more on the issues of large models solving graph tasks without addressing the generalization challenges that may arise in real-world causal reasoning scenarios. Additionally, their evaluation tasks did not include counterfactual tasks and suffered from a limited number of problems with relatively low complexity.

\vspace{0.2cm}
\noindent \textbf{Generalization of LLMs.}
Generalization refers to the model's ability to perform well on new, unseen data that wasn't part of its training set. A well-generalized model can accurately predict outcomes or classify data points beyond the examples it was trained on \citep{barbiero2020modelinggeneralizationmachinelearning}. Many evaluation research on the generalization ability of LLMs, such as graph reasoning \citep{zhang2024llmgraphreasoninggeneralize}, input length \citep{NEURIPS2022_fb7451e4}, temporal generalization \citep{zhu2024llmoutdatedevaluatingllms} and others, have been conducted. Besides, recently \citet{joshi2024llmspronefallaciescausal} researched to investigate the factors influencing causal inference in LLMs. Beyond existing research, there is still a lack of evaluation regarding the generalization ability of LLMs in causal inference tasks.

\section{Methodology}
\label{sec:method}
 In this section, we first introduce the principle for our benchmark dataset construction in Section~\ref{sec:method-taskselection}. Then we gave a formal definition of four causal inference tasks: Causal Path Finding (CP), Backdoor Adjustment (BA), Factual Inference (FI), and Counterfactual Inference (CI) in \ref{sec:method-taskdefinition}. After that, we will introduce a causal inference benchmark generation framework in \ref{sec:method-benchgenframe}. Finally, we will introduce the prompt design in \ref{sec:method-promptdesign}.

\subsection{Principles of Benchmark Construction}
\label{sec:method-taskselection}
To assess the generalization ability of LLMs in causal inference tasks, it is essential to construct a benchmark that ensures two key criteria: first, it must effectively evaluate the LLMs' causal inference capabilities; second, the questions posed must be entirely new and unseen. 
Therefore, we must first identify what can represent LLMs' ability to make causal inferences. 

According to \citet{10.5555/3238230}, there are three ladders in causality: observation, intervention, and counterfactual. 
Observation means determining the associations relationships between phenomena based on passive observations, but these associations may be non-causal.

Intervention means estimating the consequence of applying an action with causal knowledge. This is important when determining the population-wide causal effect. It needs to know not only the causal dependencies but also how to control confounders to split causal effects from non-causal ones.

At the top of the ladder, counterfactual means doing inference in a fictional world constructed from observed real world realities, this is a individual wide retrospect ability which needs not only the understanding of real world causal relations, but also the ability to modify the causal structure based on counterfactual assumptions.

Both intervention estimation and counterfactual inference abilities are essential for solving causal inference tasks. Therefore, the ability to address intervention prediction and counterfactual inference problems could serve as a representative measure of the capacity to solve causal inference problems. Based on this assumption, we selected four tasks in intervention and counterfactual ladders as representative of causal inference tasks: Causal Path Finding (CP) and Backdoor Adjustment (BA) in the intervention ladder, Factual Inference (FI) and Counterfactual Inference (CI)  in the counterfactual ladders. We assume that the performance of the selected four tasks across two causal ladders can represent the ability to solve causal inference task.

Once we determine how to measure causal inference ability, we can assess to what extent can they generalize. For LLMs, the ability to address the four problems above in unseen causal scenarios can serve as a representation of their generalization capacity in causal inference.

\subsection{Evaluation Task Definition}
\label{sec:method-taskdefinition}
We have discussed the principles for benchmark construction in Section~\ref{sec:method-taskselection}. Now, we start to give formal definitions for evaluation tasks. In this work, we used a causal graph as defined in Structural Causal Model theory \citep{10.5555/1642718} to encode a causal scenario.
A causal graph is a Directed Acyclic Graph (DAG), which encodes the causal relations between phenomena in a causal scenario.  A node in a causal graph represents a phenomena, and a directed edge between two nodes represents that the phenomenon of the starting node of the edge has a causal effect on the end one. Based on the causal graph, we gave a formal definition to the four tasks mentioned in Section~\ref{sec:method-taskdefinition}.
\subsubsection{Causal Path Finding (CP)}
\label{sec:method-taskdefinition-cp}
Given a causal graph $G$ and two nodes $A, Z$, the causal path finding task is to find all the paths between $A$ and $Z$ that only contain forward edges such as $A \rightarrow B \rightarrow Z$.

\subsubsection{Backdoor Adjustment (BA)}
\label{sec:method-taskdefinition-ba}
Given a causal graph $G$ and two nodes $A, Z$, 
a backdoor path is a path that starts with an edge pointed to $A$, for example, $A \leftarrow B \rightarrow C \rightarrow Z$. Such paths will cause non-causal information flow between node $A$ and node $Z$ and further influence the causal effect estimation between them. There are three types of junctions between nodes: chain junction ($A \rightarrow B \rightarrow C$), fork junction ($A \leftarrow B \rightarrow C$), and collider ($A \rightarrow B \leftarrow C$). For chain and fork junctions, we can block the path by controlling $B$. For colliders, the path is naturally blocked. Instead, conduct control on node $B$ will cause information flow between $A$ and $C$. The backdoor adjustment aims to block all backdoor paths by applying control on a set of nodes in these paths \citep{10.5555/3238230}. Besides backdoor paths, other non-causal paths start with an arrow from node $A$. Because it is non-causal, it must have at least a collider, for example $A \rightarrow B \rightarrow C \leftarrow D$, so it is naturally blocked. Solving this task requires applying appropriate control based on three junction types to block all backdoor paths and not open the information flow in other non-causal paths other than backdoor paths.

\subsubsection{Factual Inference (FI)}
\label{sec:method-taskdefinition-fi}
Based on the edges between nodes in a causal graph $G$, node function sets $F$ describe the causal dependencies between variables. In this work, we constrain the functions in $F$ as boolean functions, i.e., if there exist paths like $A\rightarrow Z$, $B\rightarrow Z$ and $C\rightarrow Z$, there should have a function in $F$ like $Z=A\lor(B\land \lnot\,C)$ which means event $Z$ happens if event $A$ happens or event $B$ happens and event $C$ not happen. Based on $F$, it is needed to use observed node state set $O$, for example $\{A, \lnot\,B, C\}$, to infer the node state in question node set $Q$, for example, determine whether $Z$ or $\lnot\,Z$.

\subsubsection{Counterfactual Inference (CI)}
\label{sec:method-taskdefinition-ci}
The difference between FI and CI is that besides $G, F, O$ and $Q$ provided in FI, a counterfactual assumption set $C$, for example, $\{\lnot\,A\}$, is also provided. To solve CI tasks, the node in $C$ should remove all edges pointed to it, that is, remove all the functions of this node in $F$, for example, remove functions like $A=...$ and set the node as the state assumed in $C$, i.e. $\lnot\,A$. Then, based on $G, F, C$, and $O$ infer the node state in $Q$.

\subsection{Benchmark Generation Framework}
\label{sec:method-benchgenframe}
To check LLMs' generalization ability on causal inference tasks, we proposed a benchmark generation frame work which can generate new, unseen causal inference question for CP, BA, FI and CI tasks. We started by constructing a random causal graph in Section~\ref{sec:method-benchgenframe-graph} and then assigned a random name to each node in Section~\ref{sec:method-benchgenframe-name} to ensure the constructed causal scenarios were unseen. Then, we generate questions to test the generalization ability based on the generated scenarios in Section~\ref{sec:method-benchgenframe-q}.
\subsubsection{Causal Graph Generator}
\label{sec:method-benchgenframe-graph}
To generate questions for these tasks, we must first generate a causal graph that encodes the causal relations between phenomena. We randomly generated DAGs to ensure the diversity of causal relations and that the causal structures were unseen.
To ensure the generated graph is acyclic, we separate nodes into different tiers and control it by parameter graph shape $S$, a list in which each item is the number of nodes in a tier. For example, $S=[2,\,2,\,2,\,2,\,2]$ (also marked as 2*5) means there are five tiers of nodes and two nodes in each tier. We only allow edges start from nodes in higher tier and ends at lower ones, this ensured that the node in the generated is ordered and will not form a cycle. There must be more than five tiers to ensure the versatility of paths. Then iterate $I$ times on each node, every time there is a possibility list $P$ for it to generate fork, chain, or collider junction with randomly selected other nodes. The more iterate times and higher possibilities are set, the more complex the generated graph will be.

\subsubsection{Name Generator}
\label{sec:method-benchgenframe-name}
Before we convert the causal graph into text-based causal questions, we need to assign names to the nodes in the graph. To test the abilities of LLMs to solve tasks in unseen scenarios, the names must be new to avoid being remembered by LLMs and retrieving commonsense knowledge. We assign a random string as a name to each node, such as "thepxexqaac". Besides, we also add some subject-specific meaning by appending term to the random name in two types for different tasks: plain name (random name + term), for example, "theghlfkaab decomposition," and name with change (change indicator + random name + term), for example, "decrease of ybihxaac virus". Although these names contain real existing words, after combining them with a random name, they should be an unseen name to a new thing whose properties are completely unseen; thus, they should not be related to any common knowledge. The terms contains four subjects (biology, chemistry, economics, and physics) from Wikipedia. We manually selected about 30 words per subject by frequency and added change indicators to each word. The plan name is for causal path finding and backdoor adjustment, while the name with change is for the remaining two tasks.

\subsubsection{Question Generator}
\label{sec:method-benchgenframe-q}
Based on the graph and node name prepared above, we can finally generate questions. \footnote{Examples of the questions are in Appendix~\ref{sec:apendix-qexample}.}

To evaluate the performance of LLMs when solving causal inference tasks in unseen scenarios, we declared in the question that the causal relations are derived from scientific research. Then, we describe the causal relation in a text format based on the generated causal graph and node names. 
After that, we needed to generate questions for different tasks.
For causal path finding and backdoor adjustment, we selected two different tiers in the graph, except the highest and the lowest tiers as cause and effect tiers. Then, we generated question text that described how we wanted to estimate the causal effect of nodes in the cause tier on that in effect tier and asked to find all causal paths or give backdoor adjustment node sets between each pair of them. We also introduced a parameter $ce\_d$ to control the relative distance between the two tiers, 1 means farthest and 0 means closest. We applied causal and non-causal pathfinding using DFS algorithm for these tasks, and the result served as ground truth for evaluation.
For FI and CI tasks, we selected the highest tier as the observed node state set $O$ and the lowest tier as the target inference node set $Q$. We randomly selected several nodes except those in $Q$ as the what-if node set $C$ and controlled the number of them by the parameter what-if number $ wi\_n$. We converted the boolean function into Python code and obtained ground truth through code execution.

\subsection{Prompt Design}
\label{sec:method-promptdesign}
To check LLMs performance on different prompts, in this work, we selected prompts including zero-shot (i.e., only input the question), 1 and 2-shot ICL, 0-2-shot CoT, and mistake hint prompt. We will discuss more details below.
\subsubsection{In-Context Learning}
\label{sec:method-promptdesign-icl}
Many works have revealed that LLMs can learn how to solve questions through examples using in-context learning (ICL) \citep{NEURIPS2020_1457c0d6}. We also conducted ICL for our tasks. For 1-shot ICL, we provided an example in which there is only one node in cause and effect tiers for CP and BA tasks and only one node in $O$, $Q$, and $C$ for FI and CI tasks. For 2-shot ICL, besides examples provided in 1-shot, we added a more complex one in which the node number mentioned above in different tiers or sets was 2. This setting covered the one-to-one and n-to-n causal scenarios. 
\subsubsection{Chain of Thought}
\label{sec:method-promptdesign-cot}
Chain of Thought (CoT) prompting guides LLMs to reason step-by-step through complex problems, improving their accuracy and interpretability by breaking down tasks into smaller, manageable components. We also conducted zero-shot CoT \cite{NEURIPS2022_8bb0d291}, 1 and 2-shot CoT prompt \citep{wei2022chain}. The design principle for 1 and 2-shot CoT is similar to 1 and 2-shot ICT, but it also provided them with examples telling LLMs how to solve  questions in different steps.
\subsubsection{Mistake Hint}
\label{sec:method-promptdesign-mishint}
LLMs can correct their mistake in previous answers after asking them to review. We wonder if this self-review could be done before they reach the answer. To test that, we proposed a prompt called "Mistake Hint Prompt", that is, remind the LLMs to avoid the mistake they may make before answering. For example, for CP tasks, we ask LLMs not to omit the correct path or contain the wrong or non-existing path in the answer and check carefully before reaching the answer. \footnote{More details of mistake hint prompt are in Appendix~\ref{sec:appendix-mistakehint}.}

\section{Experiment}
\label{sec:exp}
In this section, we will introduce the experiment details using the proposed benchmark generation framework and designed prompts.
\subsection{Setup}
\label{sec:exp-setup}
\textbf{Dataset Construction.} We used our framework to generate the benchmark data for the four tasks mentioned in Section~\ref{sec:method}. In this work, we give the same possibilities for generating chain, fork, and collider junction by setting $P=[0.1,\,0.1,\,0.1]$. To generate graphs with different complexities, we set graph shape $S$ to 1*5, 2*5, 1*6, and 2*6 for CP and BA, 1*5, 2*5, 1*6, 2*6 and 3*5 for FI and CI. Edge generates iterate time $I$ between 3 and 6 for all graph shapes. We generated 50 graphs per condition combination. For question generation, we set $ce\_d$ to 1 and 0.5 for CP and BA questions in 1*6 and 2*6, $wi\_n$ to 1, 2, and 3 for all IF and CF questions. Out of concern for time consumption, we filtered some questions that were too complex in CP and BA tasks. This didn't affect the overall complexity of the graph group. We used the average indegree of nodes and average chain, fork, and collider junction numbers to represent the complexity of a graph. We take the average on the four indexes over all graphs with the same graph shape to represent the overall question complexity generated based on it. Finally, we got 793 questions with $ce\_d=1$ and 395 questions with $ce\_d=0.5$ for CP and BA tasks, and 1000 questions per $wi\_n$ for FI and CI tasks. More details can be seen in Table~\ref{tab:bench_conf} and Table~\ref{tab:bench_cf}.
\begin{table}
  \centering
  \begin{tabular}{lc lc lc lc lc}
    \hline
    \textbf{GS} & \textbf{QN} & \textbf{IND} & \textbf{CH} & \textbf{FO} & \textbf{CO}\\
    \hline
    1*5     &200  &1.23  &4.88  &3.5   &3.44      \\
    1*6     &200  &1.4   &7.82  &6.02  &5.96      \\
    2*5     &198  &1.62  &17.06 &16.18 &16.12     \\
    2*6     &195  &1.72  &24.43 &21.69 &21.78     \\
    \hline
    Total   &793  &-     &-     &-     &-         \\
    \hline
  \end{tabular}
  \caption{Benchmark details for CP and BA tasks. GS: graph shape, QN: number of questions; IND: average node indegree; CH, FO, CO: average number of chain, fork, and collider junctions. The question number for 1*6 and 2*6 is the same in $ce\_d=0.5$ and $ce\_d=1$. Here only shows the number of $ce\_d=1$.}
  \label{tab:bench_conf}
\end{table}

\begin{table}
  \centering
  \begin{tabular}{lc lc lc lc lc}
    \hline
    \textbf{GS} & \textbf{QN} & \textbf{IND} & \textbf{CH} & \textbf{FO} & \textbf{CO}\\
    \hline
    1*5     &200  &1.23  &4.88  &3.5   &3.44      \\
    1*6     &200  &1.4   &7.82  &6.02  &5.96      \\
    2*5     &200  &1.63  &17.32 &16.51 &16.39     \\
    2*6     &200  &1.75  &25.4  &22.51 &22.67     \\
    3*5     &200  &1.82  &33.47 &32.55 &32.01     \\
    \hline
    Total   &1000 &-     &-     &-     &-         \\
    \hline
  \end{tabular}
  \caption{Benchmark details for FI and CI tasks. GS: graph shape, QN: number of questions; IND: average node indegree; CH, FO, CO: average number of chain, fork, and collider junctions. The question number $wi\_n=1$, $2$ or $3$ is same. Here, it only shows the number of $wi\_n=1$.}
  \label{tab:bench_cf}
\end{table}
\vspace{0.2cm}
\noindent\textbf{Models.} To make the experimental results more representative, we tested five leading LLMs on the proposed benchmark dataset, including GPT-3.5-Turbo, GPT-4o, Gemini-1.5-Pro, Claude-3-5-Sonnet-20240620 and Meta-Llama-3.1-405B-Instruct. We used default settings for all models.

\vspace{0.2cm}
\noindent\textbf{Prompt.} To evaluate the performance of LLMs under different prompts, we used prompts mentioned in Section~\ref{sec:method-promptdesign}. We applied zero-shot, 1 and 2-shot ICL, 0, 1 and 2-shot CoT, and mistake hint in Section~\ref{sec:exp-result-gsp} and \ref{sec:exp-result-tsc} and zero-shot in Section~\ref{sec:exp-result-nt}.

\vspace{0.2cm}
\noindent\textbf{Metrics.} We used accuracy as  evaluation metric. Answers from LLMs were first extracted by GPT-4o and then assessed with exact-match scoring.

\begin{figure}
    \centering
    \includegraphics[width=1\linewidth]{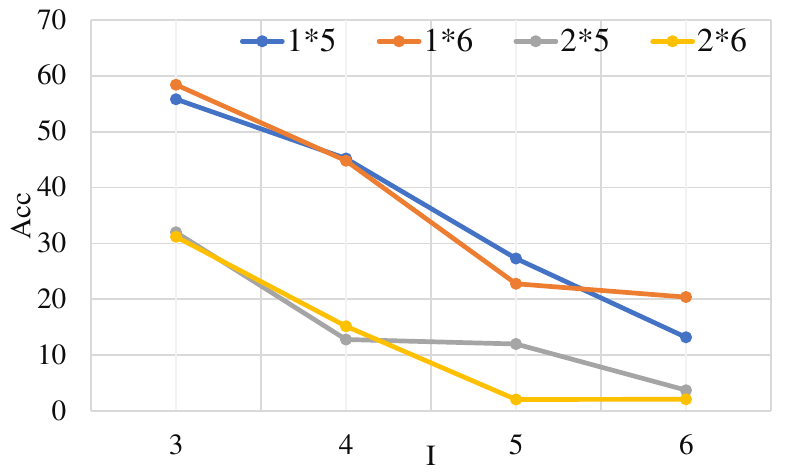}
    \caption{Accuracy of GPT-4o using zero-shot prompt on different edge iteration times for BA.}
    \label{fig:res-ba-i}
\end{figure}

\begin{table*}
  \centering
  \resizebox{\textwidth}{!}{
  \begin{tabular}{l|l|lllll|lllll}
    \hline
    \multirow{2}{*}{\textbf{Model}} & \multirow{2}{*}{\textbf{Prompt}} & \multicolumn{5}{c|}{\textbf{CP}} & \multicolumn{5}{c}{\textbf{BA}} \\ \cline{3-12}
    & & 1*5 & 1*6 & 2*5 & 2*6 & Avg & 1*5 & 1*6 & 2*5 & 2*6 & Avg\\
    \hline
 \multirow{7}{*}{GPT-4o} & 0-shot & 67.27 & 54.65 & 32.53 & 17.74 & 43.22 & 35.37 & 36.6 & 15.25 & 12.92 & 25.13\\
 & 1-shot & 82 & 70 & 40.4 & 18.46 & 52.96 & 34.5 & 39 & 16.67 & 10.26 & 25.22\\
 & 2-shot & 85 & 72.5 & 40.1 & 15.9 & 53.66 & 36.5 & 35.5 & 16.67 & 9.23 & 24.59\\
 & 0-CoT & 76 & 61 & 30.96 & 19.49 & 47.1 & 35 & 38 & 17.68 & 14.36 & 26.36\\
 & 1-CoT & 88 & 80.5 & 41.41 & 25.64 & 59.14 & 35.68 & 39 & 16.67 & 13.33 & 26.26\\
 & 2-CoT & 92.5 & 77 & 44.95 & 23.59 & \textbf{59.77} & 36.5 & 40.5 & 16.67 & 11.79 & \textbf{26.48}\\
 & mis-hint & 69 & 66.5 & 31.31 & 22.05 & 47.41 & 34 & 36 & 15.15 & 16.92 & 25.6\\
 \hline
  \end{tabular}
  }
  \caption{\label{tab:res-cpba-pgs-gpt-4o}
    Accuracy of GPT-4o using various prompts on different difficulty levels of CP and BA tasks.
  }
\end{table*}
\begin{table*}
  \centering
  \resizebox{\textwidth}{!}{
  \begin{tabular}{l|l|llllll|llllll}
    \hline
    \multirow{2}{*}{\textbf{Model}} & \multirow{2}{*}{\textbf{Prompt}} & \multicolumn{6}{c|}{\textbf{FI}} & \multicolumn{6}{c}{\textbf{CI}} \\ \cline{3-14}
    & & 1*5 & 1*6 & 2*5 & 2*6 & 3*5 & Avg & 1*5 & 1*6 & 2*5 & 2*6 & 3*5 & Avg\\
    \hline
    \multirow{7}{*}{GPT-4o} & 0-shot & 64.05 & 62.76 & 59.01 & 52.58 & 44.97 & 56.7 & 33.13 & 33.7 & 41.71 & 41.11 & 34.99 & 36.98\\
 & 1-shot & 78.17 & 79.5 & 79 & 69.5 & 65.5 & 74.32 & 42.55 & 38.67 & 66.15 & 63.64 & 62.98 & 55.56\\
 & 2-shot & 73.23 & 70.5 & 72 & 73.5 & 73 & 72.44 & 39.66 & 39.54 & 63.54 & 61.18 & 57.45 & 52.86\\
 & 0-CoT & 81.22 & 78.5 & 76.5 & 70 & 65 & 74.22 & 38.68 & 37.5 & 61.21 & 65.13 & 66.72 & 54.92\\
 & 1-CoT & 83.84 & 81 & 78.79 & 73 & 64.47 & 76.23 & 36.58 & 43.81 & 67.47 & 63.31 & 61.5 & 55.27\\
 & 2-CoT & 71.86 & 71.5 & 69.85 & 62 & 61.31 & 67.3 & 43.89 & 42.04 & 65.52 & 69.11 & 60.41 & \textbf{56.68}\\
 & mis-hint & 85.43 & 81 & 80 & 69 & 75.76 & \textbf{78.23} & 39.73 & 38.58 & 66.38 & 65.31 & 62.48 & 55.32\\

    \hline
  \end{tabular}
  }
  \caption{\label{tab:res-fici-pgs-gpt4o}
    Accuracy of GPT-4o using various prompts on different difficulty levels of FI and CI tasks.
  }
\end{table*}

\subsection{Are LLMs capable of handling causal inference tasks?}
\label{sec:exp-result-gsp}
To evaluate the LLMs' generalization abilities for causal inference, we first tested their performance on four tasks using different prompts and assessed their ability to solve problems of varying difficulty. The accuracy of CP and BA only contains questions with $ce\_d=1$. Experiment results of GPT-4o are shown in Table~\ref{tab:res-cpba-pgs-gpt-4o} and \ref{tab:res-fici-pgs-gpt4o}.\footnote{The complete experiment results are provided in the Appendix~\ref{sec:apendix-expres-pgs}.}

From the result, we can see in the CP task, for most models, CoT showed the best performance, and cloud-3.5-sonnet achieved the best performance using one-shot-CoT with an average accuracy of 88.57\%. Although performing well on solving simple CP questions, LLMs faced a performance drop when the question complexity increased. In some cases, the performance gap can be extremely large. For example, the accuracy of GPT-4o using 2-shot CoT is 92.5\% on 1*5 and 23.59\% on 2*6 questions, with an accuracy gap of 68.91\%. This phenomenon indicates that only conducting evaluations on small-scale questions may overestimate LLM's performance.

In the BA task, LLMs seem to struggle at solving this task as all LMs reached average performance between 20\% and 30\%, and there are no noticeable performance differences among different prompts. We observed that sometimes there was a performance increase with the increase in graph complexity. We conducted a deeper investigation into this phenomenon by checking the accuracy over different edge iteration time $I$. We took GPT-4o using zero-shot as a case study. As shown in Figure~\ref{fig:res-ba-i}, with the increase of $i$ within the same graph shape, the performance dropped. This suggests the complexity of the problem still influences that model performance. This may be because although there is one more node in 1*6 than 1*5 when $ce\_d=1$, there is still only one tier before the cause tier. Because edges are constructed with a fixed probability of randomly selected lower-level nodes, the additional lower-level nodes reduce the likelihood of the highest-level node pointing to the node in the causal layer, decreasing the probability of forming a backdoor path. So, there may be a performance increase in BA tasks between 1*5 and 1*6.

In the FI task, we also observed a performance drop with an increased graph shape. Both ICL and CoT prompts showed great performance improvement. Llama-3.1-405B showed the best performance on this task using two-shot-ICL with an average accuracy of 83.35\%. GPT-4o and Llama-3.1-405B have shown relatively stable performance across problems of varying complexity.

In the CI task, All models, except for GPT-3.5-turbo, achieved an average accuracy of 50-60\%. Llama-3.1-405B with one-shot ICL got the highest 60.86\% accuracy on this task.

\begin{figure}
    \centering
    \includegraphics[width=1\linewidth]{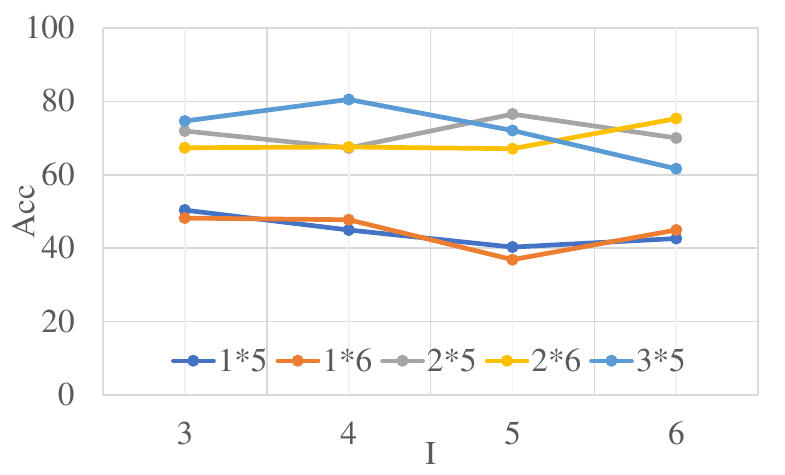}
    \caption{Accuracy of Llama-3.1-405B using one-shot ICL on different edge iteration times for CI.}
    \label{fig:res-ci-i}
\end{figure}

While GPT-3.5-Turbo showed a performance drop with increased complexity, other LLMs showed a reverse trend that reached better performance on more complex graphs. We also conducted a research on different edge iteration times as in BA. We take Llama-3.1-405B as a case study. As shown in Figure~\ref{fig:res-ci-i}, the model seems not sensitive to the question complexity of this task. We will leave the investigation of the reasons behind this phenomenon for future work.

\subsection{Is Mistake Hint Prompt Effective?}
\label{sec:exp-result-mishint}
In Section~\ref{sec:method-promptdesign-mishint}, we proposed a mistake hint prompt, that is, to remind the model to avoid making certain kinds of mistakes in their answers. But is this prompt effective?
In the experiment result in Section~\ref{sec:exp-result-gsp}, in the CP task, the mistake hint prompt was slightly better than the raw input but was still left behind ICL and CoT. In the BA task, there was no obvious difference between all prompts. In the FI task, the mistake hint prompt showed the best performance among all prompts on GPT-4o and competitive performance on GPT-3.5-Turbo and LLaMa-3.1-405B. The CI task showed the best performance among all prompts on GPT-3.5-Turbo and competitive performance on GPT-4o and LLaMa-3.1-405B. Since mistake hint has a shorter length than a few shot ICL and CoT and a near-top performance, it could be a considerable prompt candidate in FI and CI tasks.
\begin{figure}
    \centering
    \includegraphics[width=1\linewidth]{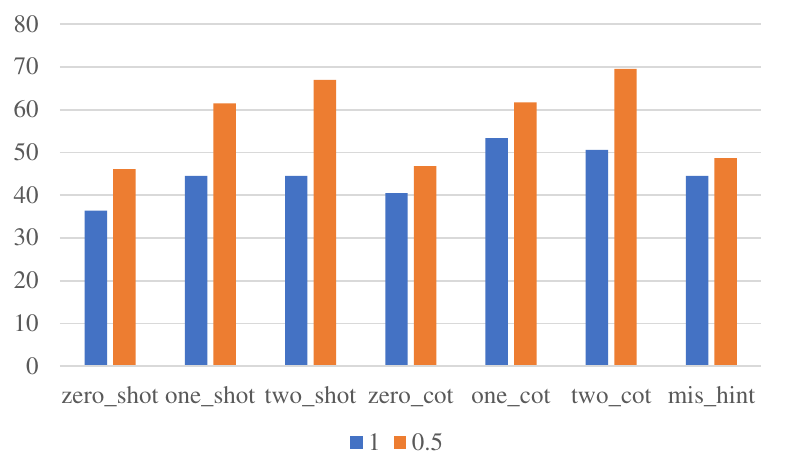}
    \caption{Accuracy of GPT-4o on different $ce\_d$ for CP.}
    \label{fig:res-cp-ced}
\end{figure}
\begin{figure}
    \centering
    \includegraphics[width=1\linewidth]{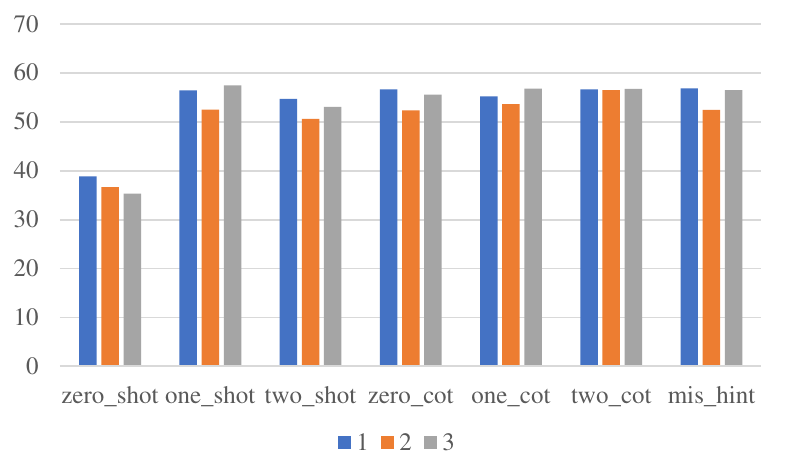}
    \caption{Accuracy of GPT-4o on different $wi\_n$ for CI.}
    \label{fig:res-ci-win}
\end{figure}

\subsection{Performance on Different Task Specific Complexity}
\label{sec:exp-result-tsc}
Besides question complexity defined by graph complexity, we also evaluated LLMs' performance on task specific complexities by controlling different parameters $ce\_d$ and $wi\_n$ introduced in Section~\ref{sec:method-benchgenframe-q}. We used average accuracy across all tested graph shapes as accuracy. For CP and BA tasks, we test different $ce\_d$ 1 and 0.5 on graph shapes 1*6 and 2*6. Figure~\ref{fig:res-cp-ced} shows the accuracy of GPT-4o on different $ce\_d$ for CP task.\footnote{The result for BA task and other models can be seen in Appendix~\ref{sec:apendix-expres-tsc}}
 The experiment result revealed that shortening the distance between the cause and effect tier increased the performance of almost all models under every prompt. In BA and CP tasks, shortening the distance between cause and effect reduces the complexity of the problems.
For CI tasks, we test different $wi\_n$ 1, 2, and 3 on all graph shapes. Figure~\ref{fig:res-ci-win} shows the accuracy of GPT-4o on different $wi\_n$ for CI task. The experimental results showed that all LLMs didn't differ much while changing the number of what-if events. This indicated that LLMs may have a robust ability to change the causal structure based on counterfactual conditions.
\begin{figure}
    \centering
    \includegraphics[width=1\linewidth]{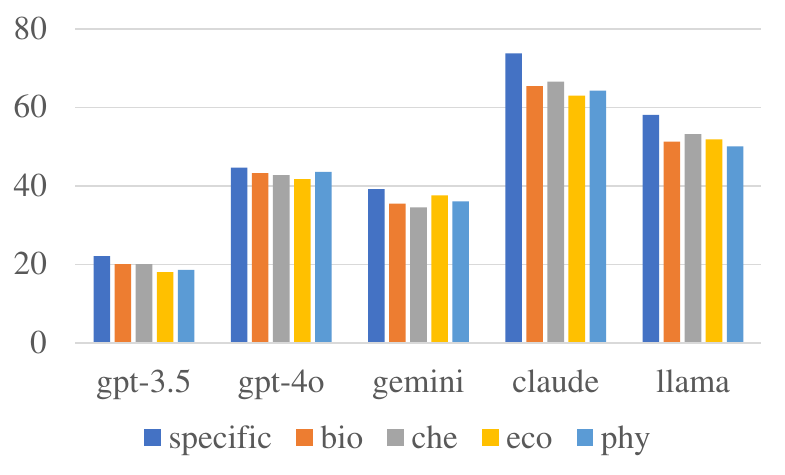}
    \caption{Accuracy on different name type in CP task.}
    \label{fig:res-nt-cp}
\end{figure}
\subsection{Can LLMs Assist in Real-world Research Efforts?}
\label{sec:exp-result-nt}
To test how well LLMs can help researchers solve new research questions, besides assigning plain names, we added subject-specific terms to each word, as mentioned in Section~\ref{sec:method-benchgenframe-name}.  We conducted a test on the whole benchmark dataset. Performance on different name types in CP task are shown in Figure~\ref{fig:res-nt-cp}.\footnote{Experiment result on other tasks can be found in Appendix~\ref{sec:apendix-expres-nt}} Experiment result reveals that except the even performance in BA task, almost all the models in remaining three tasks showed performance drop when added with actual existing words. This indicated that LLMs may have difficulty understanding how to solve BA tasks. They tend to use commonsense knowledge when real existing terms appear, which may lead them to draw incorrect conclusions when faced with new phenomena.

\section{Conclusion}
\label{sec:conclusion}
In this work, we proposed a benchmark generation framework that can generate new, unseen causal questions to test the generalization ability of LLMs on causal inference on four tasks: causal path finding (CP), backdoor adjustment (BA), factual inference (FI) and counterfactual inference (CI). We conducted extensive experiments based on the benchmark dataset generated by the proposed framework. Our experiments reveal that while LLMs are good at solving simple CP, FI and complex CI tasks, they encounter difficulties when solving BA tasks. Even when faced with new data, they tend to rely on common knowledge to solve problems, which may lead to errors in research involving novel phenomena. Overall, they exhibited limited and inconsistent generalization performance on causal inference tasks.

\section{Limitation}
\label{sec:Limitation}
Despite the contributions made in this work, there still are some limitations. Although our benchmark generation framework is a universal framework capable of generating a comprehensive amount of test data, due to time and budget constraints, we only generated and evaluated data with a limited amount and diversity. Besides, the reason for better performance on more complex CI questions shown by LLMs still needs to be answered. In addition, the four tasks we selected 
they may not fully, but at least partially represent the abilities of the models for causal inference. How to construct a comprehensive, complete causal inference evaluation method that does not rely on common knowledge still requires further research. Our work is limited to English, and performance in other languages needs further exploration.


\bibliography{anthology,custom}

\appendix
\newpage
\section{Question Examples}
\label{sec:apendix-qexample}
\begin{figure*}
    \centering
    \includegraphics[width=1\linewidth]{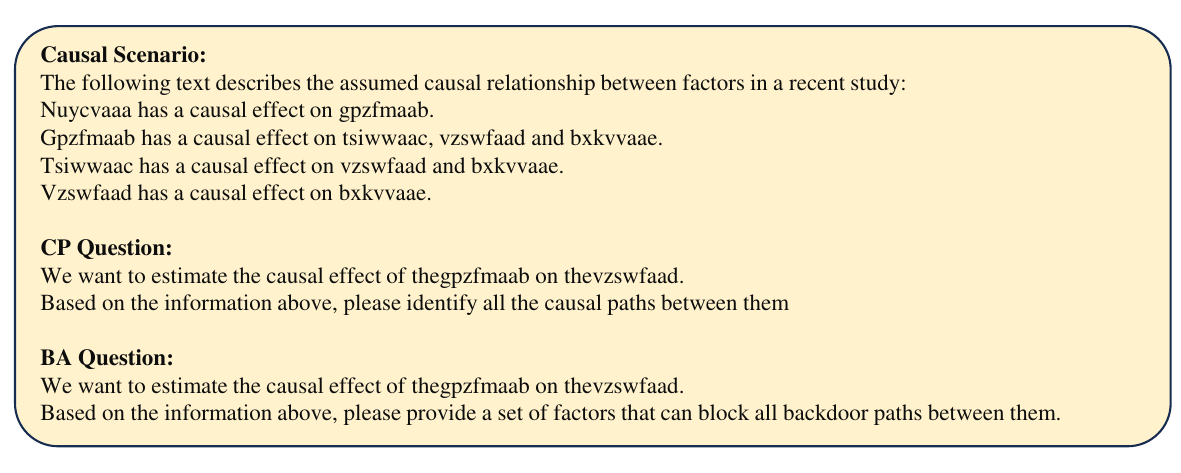}
    \caption{Example of question of CP and BA task assined with only random name.}
    \label{fig:example-cpba-q}
\end{figure*}
\begin{figure*}
    \centering
    \includegraphics[width=1\linewidth]{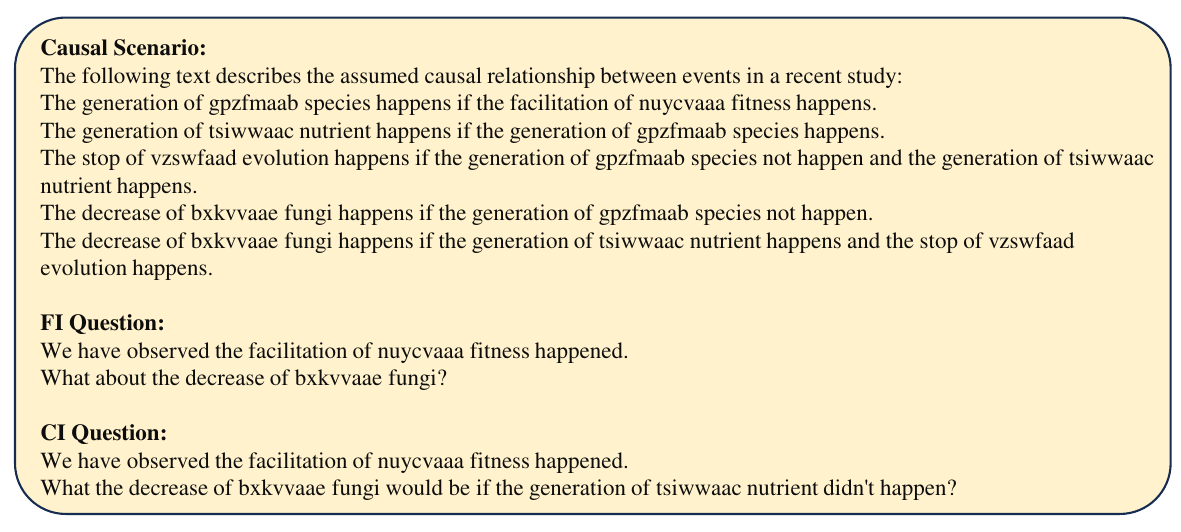}
    \caption{Example of question of FI and CI task assined with biology terms.}
    \label{fig:example-fici-q}
\end{figure*}
Figure~\ref{fig:example-cpba-q} and \ref{fig:example-fici-q} show examples of questions for CP, BA and FI, CI tasks assigned with different type of names.
\section{Mistake Hint Prompt}
\label{sec:appendix-mistakehint}
\begin{figure*}
    \centering
    \includegraphics[width=1\linewidth]{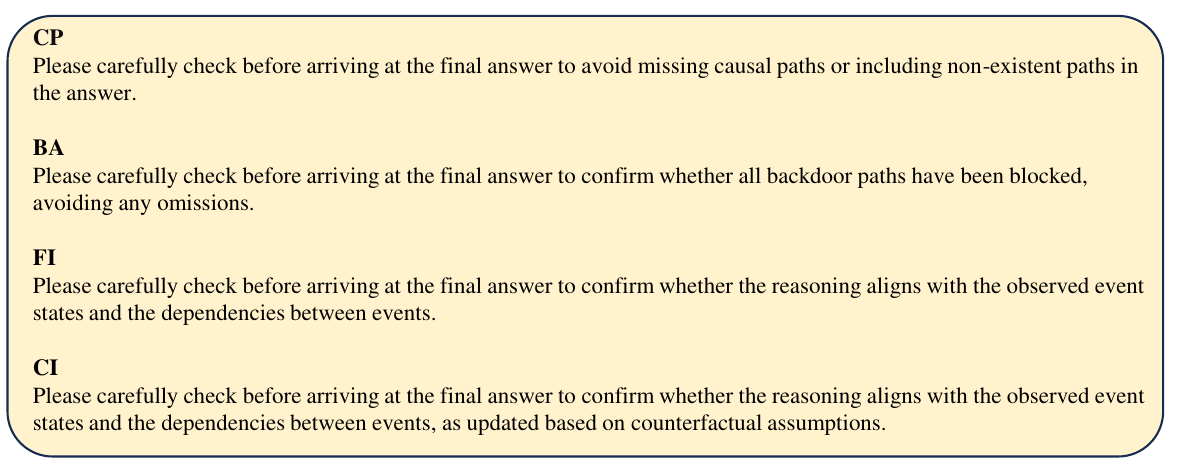}
    \caption{Details for mistake hint prompt.}
    \label{fig:example-mishint}
\end{figure*}
Figure~\ref{fig:example-mishint} shows mistake hint prompt for four different tasks.

\section{Experiment Results}

\label{sec:apendix-expres}
\subsection{Experiment Result On Different Question Complexities and Prompts}
\label{sec:apendix-expres-pgs}
\begin{table*}
  \centering
  \resizebox{\textwidth}{!}{
  \begin{tabular}{l|l|lllll|lllll}
    \hline
    \multirow{2}{*}{\textbf{Model}} & \multirow{2}{*}{\textbf{Prompt}} & \multicolumn{5}{c|}{\textbf{CP}} & \multicolumn{5}{c}{\textbf{BA}} \\ \cline{3-12}
    & & 1*5 & 1*6 & 2*5 & 2*6 & Avg & 1*5 & 1*6 & 2*5 & 2*6 & Avg\\
    \hline
 \multirow{7}{*}{gpt-3.5-turbo} & 0-shot & 42.79 & 27.33 & 4.76 & 3.79 & 19.8 & 32.73 & 37.3 & 16.68 & 11.9 & 24.75\\
 & 1-shot & 54 & 37.5 & 9.6 & 5.13 & 26.73 & 35.68 & 40.5 & 15.15 & 14.36 & 26.52\\
 & 2-shot & 53.5 & 39.5 & 8.59 & 6.67 & 27.24 & 37.5 & 42.5 & 16.67 & 11.79 & 27.24\\
 & 0-CoT & 32 & 26 & 4.59 & 5.64 & 17.19 & 33 & 37 & 14.65 & 14.87 & 24.97\\
 & 1-CoT & 60.5 & 43.5 & 4.06 & 5.13 & 28.54 & 34.5 & 38 & 13.13 & 10.77 & 24.21\\
 & 2-CoT & 59 & 40.7 & 8.12 & 10.77 & \textbf{29.84} & 39 & 43.5 & 16.67 & 13.92 & \textbf{28.41}\\
 & mis-hint & 48 & 31.5 & 5.61 & 1.54 & 21.87 & 35 & 33 & 18.69 & 13.33 & 25.09\\
 \hline
 \multirow{7}{*}{GPT-4o} & 0-shot & 67.27 & 54.65 & 32.53 & 17.74 & 43.22 & 35.37 & 36.6 & 15.25 & 12.92 & 25.13\\
 & 1-shot & 82 & 70 & 40.4 & 18.46 & 52.96 & 34.5 & 39 & 16.67 & 10.26 & 25.22\\
 & 2-shot & 85 & 72.5 & 40.1 & 15.9 & 53.66 & 36.5 & 35.5 & 16.67 & 9.23 & 24.59\\
 & 0-CoT & 76 & 61 & 30.96 & 19.49 & 47.1 & 35 & 38 & 17.68 & 14.36 & 26.36\\
 & 1-CoT & 88 & 80.5 & 41.41 & 25.64 & 59.14 & 35.68 & 39 & 16.67 & 13.33 & 26.26\\
 & 2-CoT & 92.5 & 77 & 44.95 & 23.59 & \textbf{59.77} & 36.5 & 40.5 & 16.67 & 11.79 & \textbf{26.48}\\
 & mis-hint & 69 & 66.5 & 31.31 & 22.05 & 47.41 & 34 & 36 & 15.15 & 16.92 & 25.6\\
 \hline
 \multirow{7}{*}{Gemini-1.5-Pro} & 0-shot & 49.25 & 52.01 & 27.58 & 17.04 & 36.6 & 33.57 & 38.74 & 13.33 & 13.74 & 24.94\\
 & 1-shot & 63 & 61 & 24.75 & 20.43 & 42.73 & 35.68 & 41 & 16.84 & 13.33 & 26.84\\
 & 2-shot & 61 & 58.5 & 29.29 & 20.62 & 42.55 & 36.68 & 38 & 18.27 & 14.36 & 26.93\\
 & 0-CoT & 65 & 59.5 & 32.32 & 24.62 & 45.52 & 34.85 & 39.5 & 18.69 & 14.36 & 26.93\\
 & 1-CoT & 63.5 & 57 & 28.28 & 20.51 & 42.5 & 35 & 41.5 & 18.18 & 13.85 & \textbf{27.24}\\
 & 2-CoT & 67.5 & 60 & 40.1 & 24.62 & \textbf{48.23} & 36 & 38.89 & 15.74 & 13.85 & 26.2\\
 & mis-hint & 51.5 & 54 & 27.78 & 16.49 & 37.63 & 32.66 & 40.2 & 9.6 & 10.26 & 23.26\\
 \hline
 \multirow{7}{*}{Claude-3.5-Sonnet} & 0-shot & 81.1 & 69.37 & 66.77 & 48.82 & 66.62 & 37.7 & 40.7 & 15.45 & 12.1 & 26.61\\
 & 1-shot & 96.5 & 90.5 & 82.83 & 66.67 & 84.24 & 36.5 & 41 & 17.17 & 11.28 & 26.61\\
 & 2-shot & 98 & 83.5 & 79.7 & 60 & 80.43 & 38.5 & 39.5 & 16.67 & 10.77 & 26.48\\
 & 0-CoT & 96 & 82.5 & 75.76 & 61.86 & 79.17 & 37.5 & 37.5 & 15.15 & 12.31 & 25.73\\
 & 1-CoT & 98.5 & 91 & 83.33 & 69.74 & \textbf{85.75} & 38.5 & 42 & 16.16 & 12.31 & 27.36\\
 & 2-CoT & 99 & 84.5 & 85.86 & 61.54 & 82.85 & 39.5 & 43.5 & 18.69 & 12.82 & \textbf{28.75}\\
 & mis-hint & 89.5 & 75 & 75.76 & 59.49 & 75.03 & 36 & 40 & 13.64 & 10.77 & 25.22\\
 \hline
 \multirow{7}{*}{Llama-3.1-405B} & 0-shot & 68.6 & 73.17 & 41.76 & 27.52 & 52.93 & 33.33 & 35.24 & 15.45 & 11.79 & 24.05\\
 & 1-shot & 94.5 & 87 & 58.88 & 38.97 & 70.08 & 36.68 & 39 & 17.68 & 10.26 & 26.01\\
 & 2-shot & 94 & 81 & 70.71 & 44.62 & 72.76 & 40 & 39.9 & 14.14 & 10.26 & 26.17\\
 & 0-CoT & 87.5 & 83.5 & 59.09 & 38.46 & 67.34 & 35.5 & 36.68 & 14.65 & 12.31 & 24.87\\
 & 1-CoT & 93.5 & 85 & 65.66 & 47.18 & \textbf{73.01} & 40.5 & 40.5 & 15.66 & 11.28 & 27.11\\
 & 2-CoT & 94.5 & 85 & 64.65 & 42.05 & 71.75 & 40.5 & 44 & 15.15 & 11.79 & \textbf{27.99}\\
 & mis-hint & 80 & 79 & 49.49 & 34.87 & 61.03 & 34.5 & 35.5 & 14.65 & 12.31 & 24.34\\

    \hline
  \end{tabular}
  }
  \caption{\label{tab:res-cpba-pgs}
    Accuracy of different models using various prompts on different difficulty levels of CP and BA tasks.
  }
\end{table*}
\begin{table*}
  \centering
  \resizebox{\textwidth}{!}{
  \begin{tabular}{l|l|llllll|llllll}
    \hline
    \multirow{2}{*}{\textbf{Model}} & \multirow{2}{*}{\textbf{Prompt}} & \multicolumn{6}{c|}{\textbf{FI}} & \multicolumn{6}{c}{\textbf{CI}} \\ \cline{3-14}
    & & 1*5 & 1*6 & 2*5 & 2*6 & 3*5 & Avg & 1*5 & 1*6 & 2*5 & 2*6 & 3*5 & Avg\\
    \hline
    \multirow{7}{*}{GPT-3.5-Trubo} & 0-shot & 51.52 & 47.01 & 28.79 & 23.93 & 15.92 & 33.34 & 25.11 & 27.2 & 21.66 & 23.1 & 13.3 & 22.01\\
     & 1-shot & 57.54 & 49.08 & 33.84 & 27.33 & 14.57 & 35.78 & 26.86 & 29.65 & 22.11 & 23.89 & 12.5 & 22.65\\
      & 2-shot & 61.9 & 54.71 & 33.84 & 26.46 & 13.07 & \textbf{37.35} & 28.62 & 31.19 & 23.37 & 22.22 & 11.52 & 23.07\\
      & 0-CoT & 56.12 & 52.88 & 23.08 & 30.21 & 20.2 & 36.42 & 31.38 & 28.98 & 24.57 & 22.16 & 13.21 & \textbf{23.82}\\
      & 1-CoT & 48.22 & 39.04 & 22.4 & 25.41 & 20.53 & 31.23 & 26.64 & 25.85 & 18.54 & 22.18 & 11.45 & 20.76\\
      & 2-CoT & 46.43 & 29.41 & 17.68 & 17.28 & 15.9 & 25.34 & 20.93 & 22.4 & 18.17 & 15.83 & 10.75 & 17.53\\
      & mis-hint & 47.72 & 45.26 & 31.31 & 36.65 & 18.88 & 35.91 & 32.48 & 33.39 & 23.57 & 25.21 & 14.02 & 25.5\\
    \hline
    \multirow{7}{*}{GPT-4o} & 0-shot & 64.05 & 62.76 & 59.01 & 52.58 & 44.97 & 56.7 & 33.13 & 33.7 & 41.71 & 41.11 & 34.99 & 36.98\\
 & 1-shot & 78.17 & 79.5 & 79 & 69.5 & 65.5 & 74.32 & 42.55 & 38.67 & 66.15 & 63.64 & 62.98 & 55.56\\
 & 2-shot & 73.23 & 70.5 & 72 & 73.5 & 73 & 72.44 & 39.66 & 39.54 & 63.54 & 61.18 & 57.45 & 52.86\\
 & 0-CoT & 81.22 & 78.5 & 76.5 & 70 & 65 & 74.22 & 38.68 & 37.5 & 61.21 & 65.13 & 66.72 & 54.92\\
 & 1-CoT & 83.84 & 81 & 78.79 & 73 & 64.47 & 76.23 & 36.58 & 43.81 & 67.47 & 63.31 & 61.5 & 55.27\\
 & 2-CoT & 71.86 & 71.5 & 69.85 & 62 & 61.31 & 67.3 & 43.89 & 42.04 & 65.52 & 69.11 & 60.41 & \textbf{56.68}\\
 & mis-hint & 85.43 & 81 & 80 & 69 & 75.76 & \textbf{78.23} & 39.73 & 38.58 & 66.38 & 65.31 & 62.48 & 55.32\\
 \hline
 \multirow{7}{*}{Gemini-1.5-Pro}  & 0-shot & 54.93 & 53.08 & 41.07 & 33.85 & 24.69 & 40.84 & 29.99 & 34.19 & 33.92 & 34.85 & 24.1 & 30.82\\
 & 1-shot & 62.9 & 66.14 & 64.14 & 60.31 & 57.36 & 62.14 & 49.16 & 48.48 & 63.08 & 62.29 & 56.78 & 56.74\\
 & 2-shot & 77.96 & 64.21 & 62.81 & 64.82 & 61.5 & \textbf{66.12} & 45.38 & 46.6 & 61.25 & 57.94 & 57.66 & 54.49\\
 & 0-CoT & 72.87 & 75.66 & 55.21 & 53.12 & 50.77 & 61.4 & 48.05 & 44.31 & 42.61 & 46.73 & 48.24 & 46.03\\
 & 1-CoT & 78.24 & 63.16 & 58.16 & 63.27 & 53.06 & 63.13 & 48.3 & 56.11 & 60.93 & 65.15 & 56.61 & \textbf{58.09}\\
 & 2-CoT & 77.46 & 67.88 & 61.11 & 59.18 & 54.5 & 63.65 & 44.48 & 49.21 & 60.28 & 62.06 & 56.25 & 55.41\\
 & mis-hint & 62.64 & 57.8 & 47.12 & 39.9 & 36.41 & 48.39 & 43.54 & 39.87 & 45.64 & 41.61 & 39.89 & 42.04\\
\hline
\multirow{7}{*}{Claude-3.5-Sonnet} & 0-shot & 72.89 & 61.67 & 58.93 & 48.58 & 35.41 & 55.55 & 31.3 & 29.99 & 47.43 & 48.65 & 35.06 & 38.78\\
 & 1-shot & 80.81 & 81.31 & 72.08 & 59.28 & 46.67 & 68.13 & 42.62 & 43.68 & 62.63 & 60.31 & 54.06 & \textbf{52.82}\\
 & 2-shot & 87.37 & 81.31 & 72.5 & 60.2 & 53.27 & 70.94 & 43.07 & 40.83 & 61.2 & 60.45 & 54.55 & 52.25\\
 & 0-CoT & 84.38 & 80.2 & 77.27 & 65.31 & 51.53 & \textbf{71.71} & 43.89 & 47.41 & 57.37 & 58.71 & 49.57 & 51.58\\
 & 1-CoT & 87.37 & 79.9 & 73.5 & 54.64 & 52.53 & 69.67 & 44.3 & 46.58 & 62.5 & 60.89 & 56.22 & 54.3\\
 & 2-CoT & 89.9 & 78.39 & 71 & 58 & 49.25 & 69.28 & 46.35 & 47.56 & 60.77 & 64.93 & 53.64 & 54.8\\
 & mis-hint & 82.65 & 77.39 & 70.35 & 58.59 & 55.9 & 69 & 43.24 & 38.77 & 57.22 & 57.42 & 57 & 51.17\\
\hline
\multirow{7}{*}{Llama-3.1-405B} & 0-shot & 65.46 & 62.73 & 61.39 & 53.81 & 43.99 & 57.29 & 31.95 & 32.65 & 41.74 & 42.91 & 37.96 & 37.52\\
 & 1-shot & 87.76 & 86.5 & 88 & 75.5 & 78.5 & 83.23 & 44.51 & 44.38 & 71.43 & 69.37 & 72.3 & \textbf{60.86}\\
 & 2-shot & 89.39 & 88.5 & 83.5 & 77 & 78.39 & \textbf{83.35} & 43.97 & 42.65 & 70.99 & 73.23 & 68.78 & 60.5\\
 & 0-CoT & 86.29 & 84.77 & 84 & 79.7 & 76.65 & 82.29 & 40 & 38.22 & 61.31 & 62.24 & 69.83 & 54.76\\
 & 1-CoT & 85.93 & 81 & 74 & 70 & 66 & 75.38 & 43.09 & 40.71 & 67.52 & 68.53 & 69.17 & 58.17\\
 & 2-CoT & 84.42 & 74 & 66 & 58 & 64.32 & 69.34 & 47.69 & 45.86 & 62.48 & 62.79 & 58.92 & 55.71\\
 & mis-hint & 83.67 & 79.9 & 82.74 & 78 & 73.37 & 79.52 & 38.48 & 41.97 & 60.14 & 67.06 & 67.97 & 55.68\\
    \hline
  \end{tabular}
  }
  \caption{\label{tab:res-fici-pgs}
    Accuracy of different models using various prompts on different difficulty levels of FI and CI tasks.
  }
\end{table*}

Table~\ref{tab:res-cpba-pgs} shows the experiment result on different question complexities and prompts for CP and BA tasks. Table~\ref{tab:res-fici-pgs} is for FI and CI tasks.
\subsection{Task Specific Complexity}
\label{sec:apendix-expres-tsc}
\begin{figure*}
    \centering
    \includegraphics[width=1\linewidth]{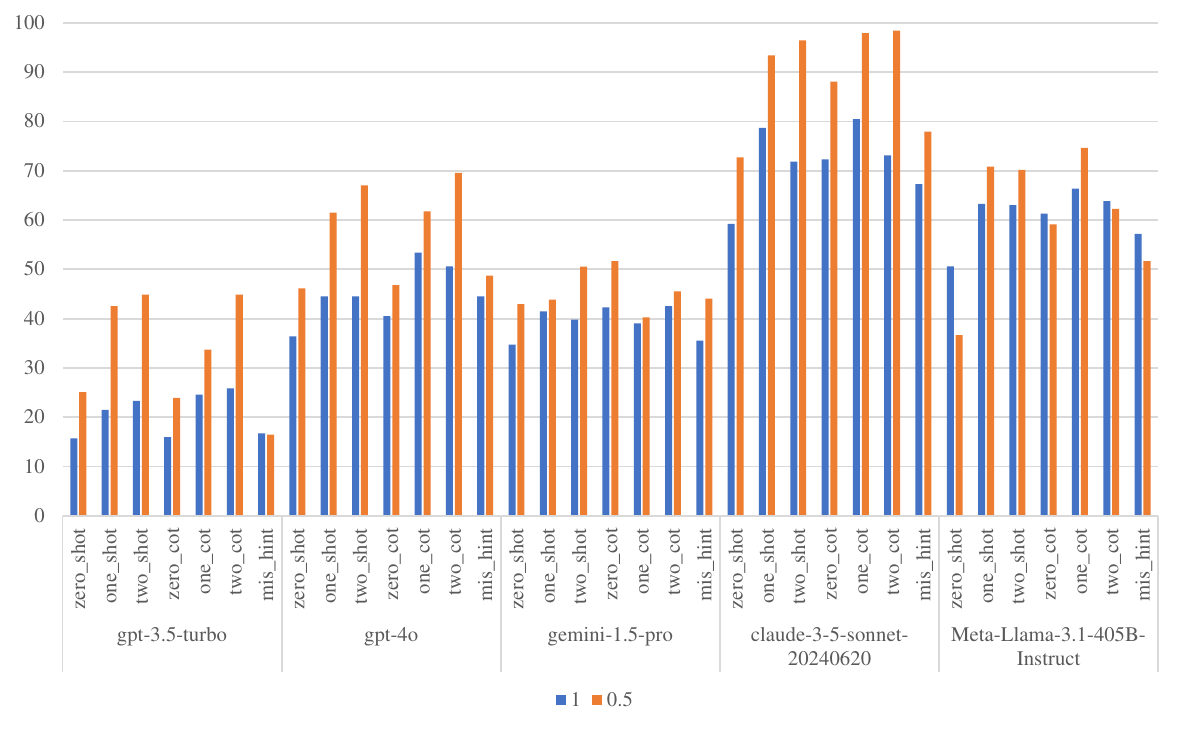}
    \caption{Accuracy on different $ce\_d$ for CP.}
    \label{fig:cp-ced}
\end{figure*}
\begin{figure*}
    \centering
    \includegraphics[width=1\linewidth]{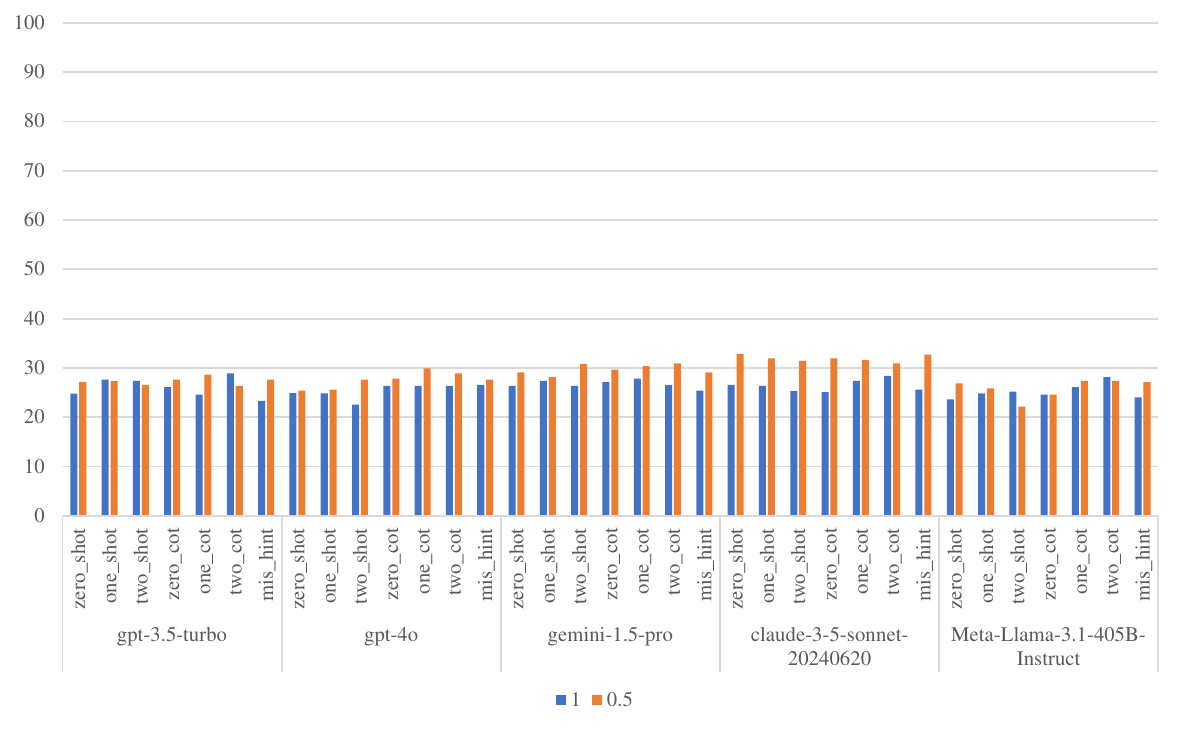}
    \caption{Accuracy on different $ce\_d$ for BA.}
    \label{fig:ba-ced}
\end{figure*}
\begin{figure*}
    \centering
    \includegraphics[width=1\linewidth]{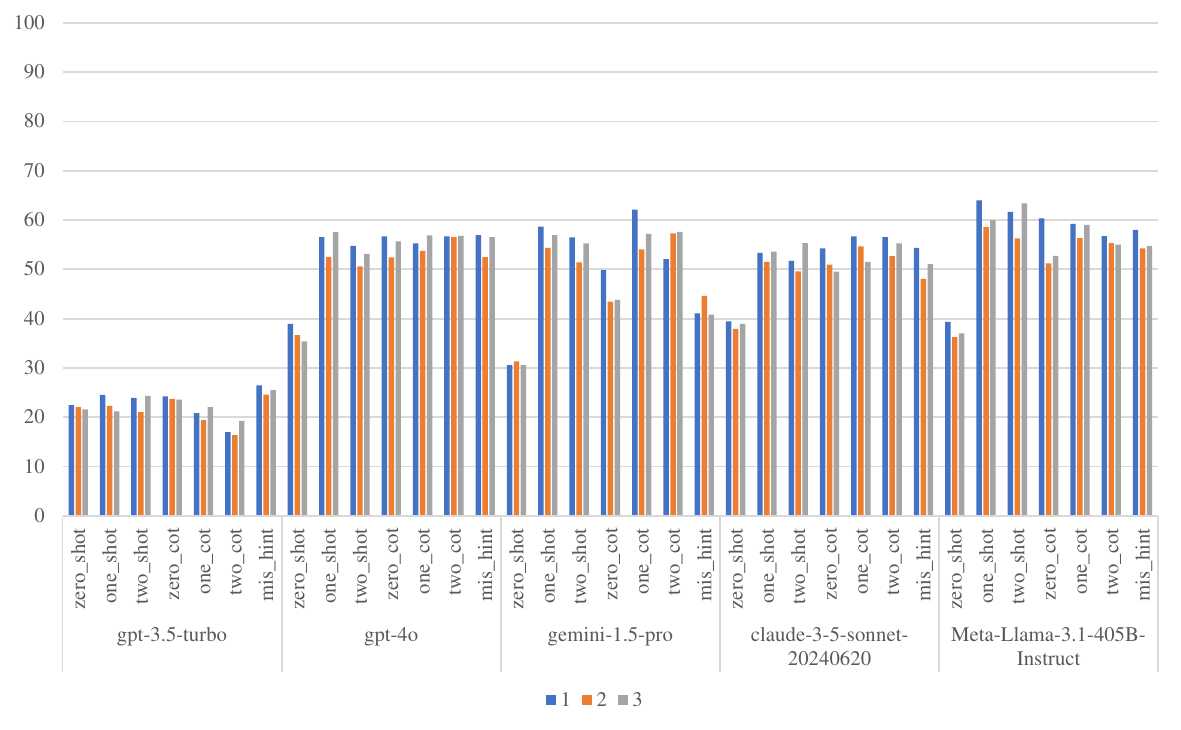}
    \caption{Accuracy on different $ce\_d$ for CI.}
    \label{fig:ci-win}
\end{figure*}
Figure~\ref{fig:cp-ced} and \ref{fig:ba-ced} shows the average accuracy with different $ce\_d$ in CP and BA tasks. Figure~\ref{fig:ci-win} shows the average accuracy with different $wi\_n$ in CI tasks.

\subsection{Performance on Different Name Type}

\begin{figure}
    \centering
    \includegraphics[width=1\linewidth]{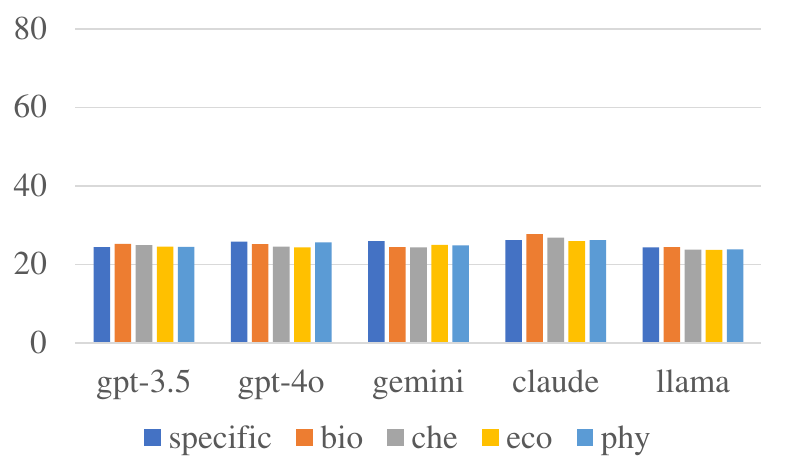}
    \caption{Accuracy on different name type in BA task.}
    \label{fig:res-nt-ba}
\end{figure}

\begin{figure}
    \centering
    \includegraphics[width=1\linewidth]{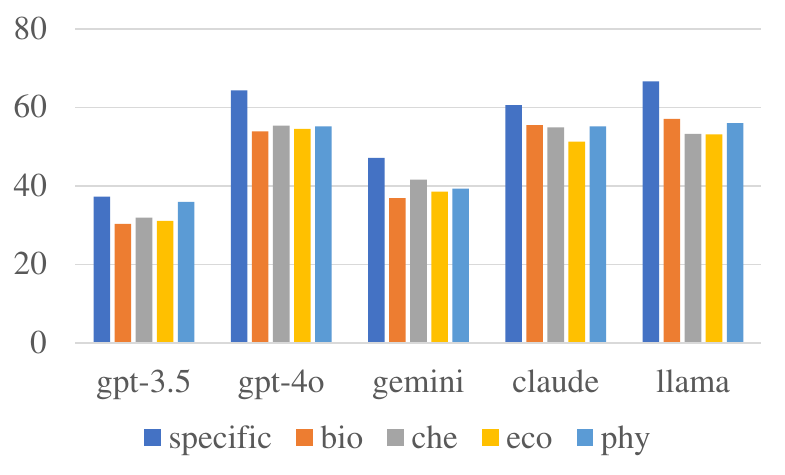}
    \caption{Accuracy on different name type in FI task.}
    \label{fig:res-nt-fi}
\end{figure}

\begin{figure}
    \centering
    \includegraphics[width=1\linewidth]{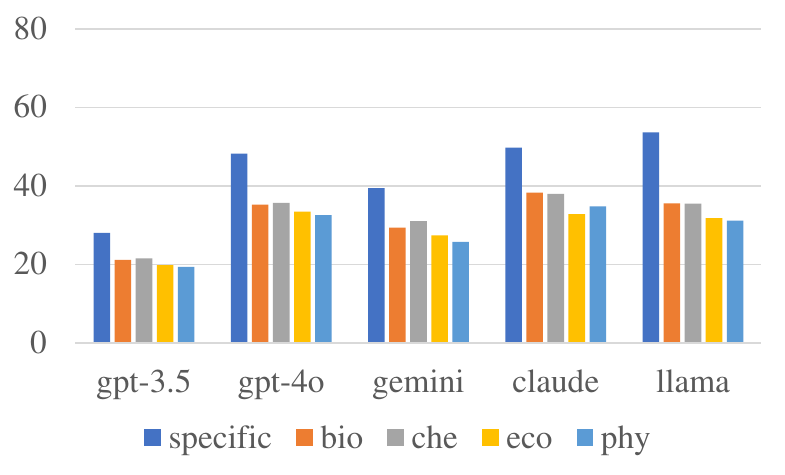}
    \caption{Accuracy on different name type in CI task.}
    \label{fig:res-nt-ci}
\end{figure}

In addition to Figure~\ref{fig:res-nt-cp}, Figure~\ref{fig:res-nt-ba}, \ref{fig:res-nt-fi} and \ref{fig:res-nt-ci} show performance while assigned different type of names in BA, FI and CI tasks.
\label{sec:apendix-expres-nt}
\end{document}